%% file: main.tex
\documentclass[conf]{new-aiaa}
\usepackage[utf8]{inputenc}

\usepackage{graphicx}
\usepackage{amsmath}
\usepackage[version=4]{mhchem}
\usepackage{siunitx}
\usepackage{longtable,tabularx}
\usepackage{subcaption}
\usepackage{float}
\usepackage[ruled,vlined]{algorithm2e}
\usepackage{siunitx}
\usepackage{gensymb}
\usepackage{epstopdf}
\usepackage{listings}
\usepackage{cleveref}

\Crefname{figure}{Fig.}{Figs.}

\title{An Open-Source Gazebo Plugin for GNSS Multipath Signal Emulation in Virtual Urban Canyons}

\author{Kartik Anand Pant\footnote{Graduate Student, School of Aeronautics and Astronautics. kpant@purdue.edu}, Zhanpeng Yang\footnote{Graduate Student, School of Aeronautics and Astronautics. yang1272@purdue.edu}, James M Goppert, \footnote{Lecturer, School of Aeronautics and Astronautics, jgoppert@purdue.edu} and Inseok Hwang \footnote{ Professor, School of Aeronautics and Astronautics, ihwang@purdue.edu and AIAA Associate Fellow }}
\affil{Purdue University, West Lafayette, IN, 47906}

\begin{document}

\maketitle

\begin{abstract} 
One of the major errors affecting GNSS signals in urban canyons is GNSS multipath error. In this work, we develop a Gazebo plugin which utilizes a ray tracing technique to account for multipath effects in a virtual urban canyon environment using virtual satellites. This software plugin balances accuracy and computational complexity to run the simulation in real-time for both software-in-the-loop (SITL) and hardware-in-the-loop (HITL) testing. We also construct a 3D virtual environment of Hong Kong and compare the results from our plugin with the GNSS data in the publicly available Urban-Nav dataset, to validate the efficacy of the proposed Gazebo Plugin. The plugin is openly available to all the researchers in the robotics community. \href{https://github.com/kpant14/multipath_sim}{\texttt{https://github.com/kpant14/multipath\_sim}}
\end{abstract}

\input{sections/01_nomenclature.tex}
\input{sections/02_introduction.tex}

\input{sections/03_background.tex}
\input{sections/04_architecture.tex}
\input{sections/05_experimental_results.tex}
\input{sections/06_conclusion.tex}

\section*{Acknowledgments}
This research is funded by the Secure Systems Research Center(SSRC) at the Technology Innovation Institute (TII), UAE. The authors are grateful to Dr. Shreekant (Ticky) Thakkar and his team members at the SSRC for their valuable comments and support.

\bibliography{bibtex}

\end{document}

%% file: sections/01_nomenclature.tex
\section{Nomenclature}

{\renewcommand\arraystretch{1.0}
\noindent\begin{longtable*}{@{}l @{\quad=\quad} l@{}}
$m$  & Multipath offset \\
$d$  & Additional distance covered by the reflected GNSS signal before reflection\\
$l$  & Additional distance covered by the reflected GNSS signal after reflection\\
$\theta_i$  & Elevation angle of the $i^{th}$ satellite\\
$\theta_0$  & Elevation angle threshold\\
$\alpha$  & Angle subtended by the normal of the reflected ray with the reflected ray\\
$P_i$  & Pseudorange of the $i^{th}$ satellite\\
$c_0$  & speed of light\\
$\Delta t$  & clock bias of the receiver with respect to the satellite\\
$e$ & measurement error\\
$\sigma$ & standard deviation of the measurement error\\
$\rho_i$  & True range of the $i^{th}$ satellite\\
$(x_i,y_i,z_i)$  & Position coordinate of the $i^{th}$ satellite\\
$(x,y,z)$  & Position coordinate of the receiver\\
$\sigma_G$ & Geometric dilution-of-precision\\
$r^{LOS}$ & Laser range for LOS ray\\
$r^{REF}$ & Laser range for Mirror ray
\end{longtable*}}

%% file: sections/02_introduction.tex
\section{Introduction}
\label{sec:intro}
\vspace{2pt}
\lettrine{T}{he} Global Navigation Satellite System (GNSS) is being widely used for numerous modern transportation and logistic applications, from school buses to military convoys. A robust GNSS positioning fix is essential for safe and secure navigation in a multitude of civil and military applications. With the advent of urban air mobility (UAM), such accurate positioning capability becomes paramount to the safe operation of aerial vehicles, whether unmanned or passenger-carrying. The accuracy of GNSS receivers is affected by several factors, for example, ionospheric errors, satellite clock errors, and multipath errors. By using differential techniques, ionospheric effects can be mitigated \cite{monteiro2005dgps}. However, the effects of multipath errors are site and time dependent and urban environments are particularly impacted by GNSS multipath. There are two major challenges for GNSS operations in an urban canyon. Firstly, the densely packed high-rise buildings significantly obstruct the line-of-sight (LOS) of satellites. Secondly, the high-rise buildings also provide ideal surfaces for the multipath of GNSS signals to occur, which further reduces the accuracy of the position fix provided by the GNSS constellation. Many studies have been conducted for multipath detection and mitigation in the past. A 3D model of the environment is constructed in \cite{Bauer3DModelling, Zhang3DModelling} to detect multipath errors by using the ray tracing method in an urban canyon environment. However, both approaches require computationally expensive ray-tracing algorithms to check every 3D surface in the virtual environment. A similar approach using ray-tracing is developed in \cite{LauGPSMultipath}, however, it requires precise knowledge of the satellite–reflector–antenna geometry, the reflector material, and antenna characteristics. Statistical methods have been proposed in \cite{LeeGPSMultipath, wu2010statistical} to detect multipath using the double difference of the pseudo-ranges and carrier phase measurements. The multipath errors have been modeled as statistical distributions in \cite{BrennerStatMultipath, WangStatMultipath, KongStatMultipath}. In \cite{SteingassMultipath, SteingassMultipathChannel}, a simulated satellite using a Zeppelin is utilized to compute the precise measurements of statistical distributions of multipath delays. Although statistical methods provide good estimates of the multipath properties, they require large-scale raw GNSS data collection to fit the error profile of a specific region and they do not accurately represent the position and time dependent offsets inherent in multipath errors.

In this work, we present an open-source GNSS sensor emulation plugin for the Gazebo simulator that can be easily integrated with popular flight control firmware such as PX4 and Ardupilot. The proposed GNSS plugin utilizes a ray-tracing algorithm on 3D building models. The building models in the Gazebo simulator are generated using data obtained from Open Street Map (OSM), to estimate the GNSS signal paths at each location in the simulated urban environment. This framework can be utilized for:
\begin{enumerate}
    \item Software-in-the-Loop (SITL) testing of UASs in a virtual urban environment constructed in Gazebo.
    \item Mixed Reality testing on real UAVs flying in the tracking volume of the motion capture (MoCap) system while a virtual copy moves in the simulated urban environment. 
    \item Computing heat maps of GNSS multipath error for particular times and locations in an urban environment.
\end{enumerate}

The main contribution of our work is to provide an open-source GNSS emulation plugin in the ROS-Gazebo environment that is accessible to the wider UAS community. The rapid adaptation of Cyber-Physical Systems (CPS) in urban environments in recent years have led to a strong emphasis on the safety and security of autonomous systems. The proposed GNSS plugin can be used as an effective tool to model and prototype the effects of GNSS multipath errors on large-scale CPS systems. It can be further utilized to identify state and time dependent vulnerabilities in the urban environment to synthesize stealthy GNSS spoofing attacks and develop effective detection and mitigation strategies against them. 

The rest of the paper is organized as follows: Section \ref{sec:math-model} describes the construction of the virtual 3D environment and the mathematical model used for the computation of pseudo-ranges in the presence of multipath errors. Section \ref{sec:arch-soft-design} highlights the architecture followed for designing the proposed GNSS plugin in Gazebo. A comparison of the performance of the proposed with a publicly available GNSS dataset of Hong Kong city is presented in Section \ref{sec:impl}. Finally, Section \ref{sec:conclusion} summarizes the research and lays out future directions.

%% file: sections/03_background.tex
\section{Background and Preliminaries}
\label{sec:math-model}
\vspace{2pt}
This section describes mathematical formulations used for modeling the multipath effects on the GNSS signals and discusses the theory of operation in estimating the receiver's position. It also describes the construction of the virtual urban 3D environment and virtual satellites in the Gazebo simulator.

The GNSS receivers periodically receive radio signals from various satellites in a constellation and compute the ranges by measuring the signal delay. These ranges are then used to estimate the position of the receiver. As the satellite and the receiver have different clocks, the receiver also estimates its own clock bias with respect to the satellite's clock. In order to uniquely calculate the coordinates of the receiver, the receiver requires a minimum of four satellites to estimate its position and the clock bias \cite{parkinson1996global}. In this work, the GNSS signals are modeled as a straight line segment between the transmitter at each satellite and the receiver on the Earth. In reality, however, these signals do not follow a straight line path due to the atmospheric effects. The straight-line assumption is valid as the effects of atmospheric errors (ionospheric and tropospheric) on GNSS signals are well modeled \cite{Klobuchar1987Ionos, Bevis1994GNSSModel}. The corrections for these errors are available on most commercially available receivers through the wide-area augmentation system (WAAS) \cite{enge1996wide}.  

This work focuses on the effects of multipath errors on GNSS localization which are dependent on the geometry of the environment. Multipath errors are difficult to model as they occur due to reflective surfaces such as tall buildings, dense vegetation, the ground, etc. In the absence of LOS to a specific satellite, the GNSS receiver may only receive reflected multipath signals. The multipath signal travels a longer distance and thus creates a synchronization bias with respect to LOS signals from other satellites, resulting in errors in code and carrier pseudo-range measurements. Ultimately, multipath errors lead to significant errors in positioning for the GNSS receiver, especially in urban canyons. 

To model the behavior of the multipath signals, first, a 3D virtual environment of a part of the city is constructed in the Gazebo simulator, followed by a ray tracing algorithm to compute the pseudo-range offset resulting from the multipath effects. The modified pseudo-ranges are then utilized to compute the position fix and the estimation accuracy, using the least squares estimation algorithm. The details of each of these steps are described below:  

\subsection{Virtual Urban Canyon and Virtual Satellites}
\begin{figure}[H]
    \centering
    \includegraphics[width=0.8\textwidth]{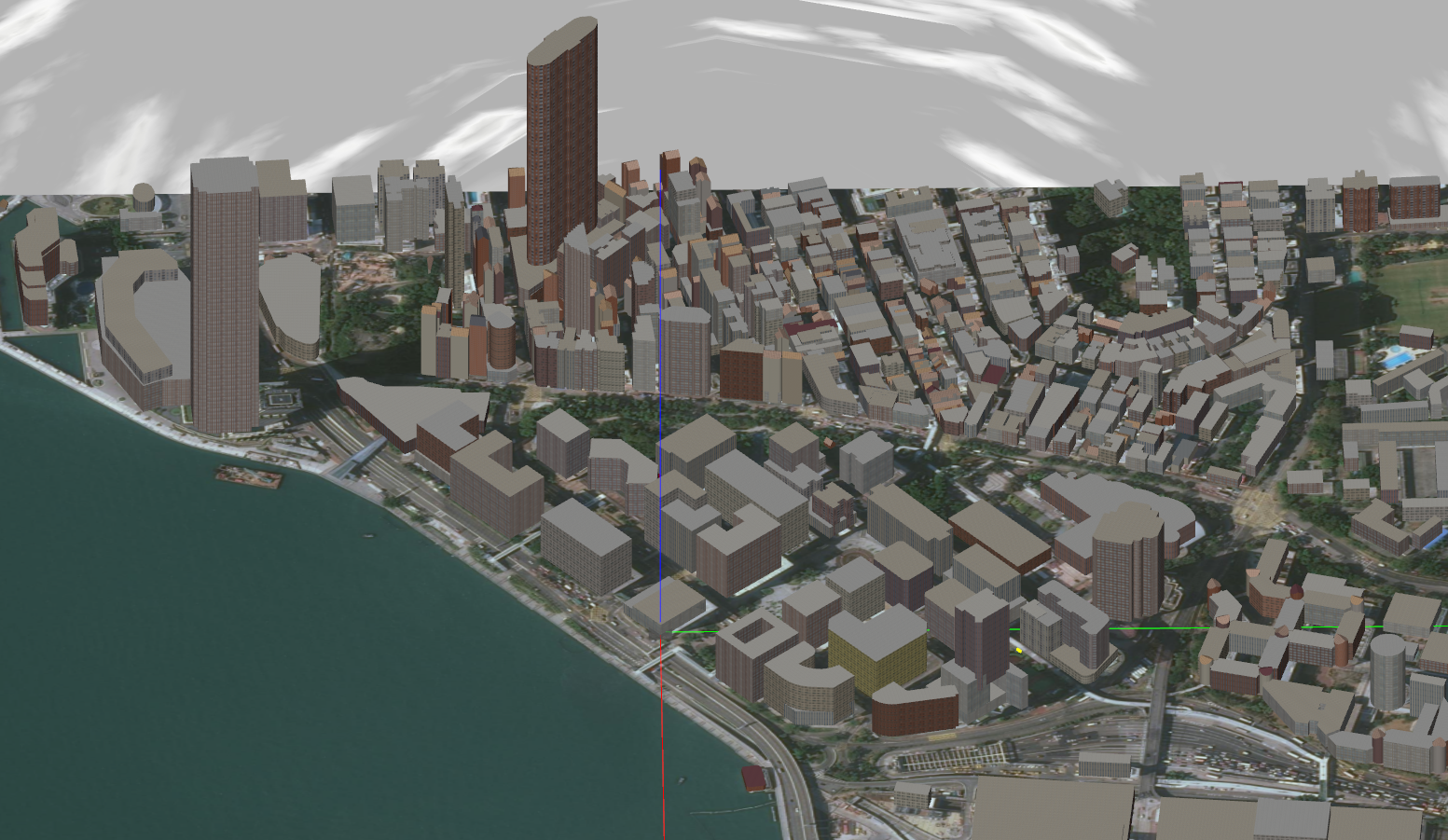}
    \caption{Virtual Hong Kong Environment in Gazebo}
    \label{fig:virtual_hk}
\end{figure}

To develop the 3D virtual environment in Gazebo, the building information for the desired part of the city is obtained from the publicly available dataset using Open Street Maps (OSM) \cite{openstreetmap}. The data from OSM is pre-processed to extract the 3D meshes and geodetic coordinates of the buildings. The geodetic positions are then converted into Universal Transverse Mercator (UTM) coordinates to align with the coordinate system of Gazebo. The virtual satellites are set up using the raw ephemeris data either retrieved from the publicly available GNSS repositories \cite{noll2015cddis} or acquiring it with an actual GNSS receiver over a period of time. The raw ephemeris data contains the Keplerian elements of the satellite's orbit, for example, orbital inclination and eccentricity. The orbital parameters are used to calculate the position of the satellites in the Earth Centered Earth Fixed (ECEF) coordinate system. The construction of the virtual 3D model of the city of Hong Kong is shown in \Cref{fig:virtual_hk}.  
\subsection{Ray Tracing Algorithm}
For each satellite, the ray-tracing algorithm simulates two rays originating from the GNSS receiver, one connecting the satellite and receiver directly and another in the complementary direction, i.e., the direction of the reflected signal. The LOS signal is simulated as a straight line segment connecting the GNSS receiver with the satellite. Similarly, the reflected signal is simulated as a straight line segment connecting the GNSS receiver and the obstruction (if there exists one). The lengths of these line segments are utilized to (1) detect the presence of multipath error and (2) calculate the multipath offset, i.e., additional distance covered by the reflected signal before reaching the receiver. 
\begin{figure}[ht]
    \centering
    \includegraphics[width=\textwidth]{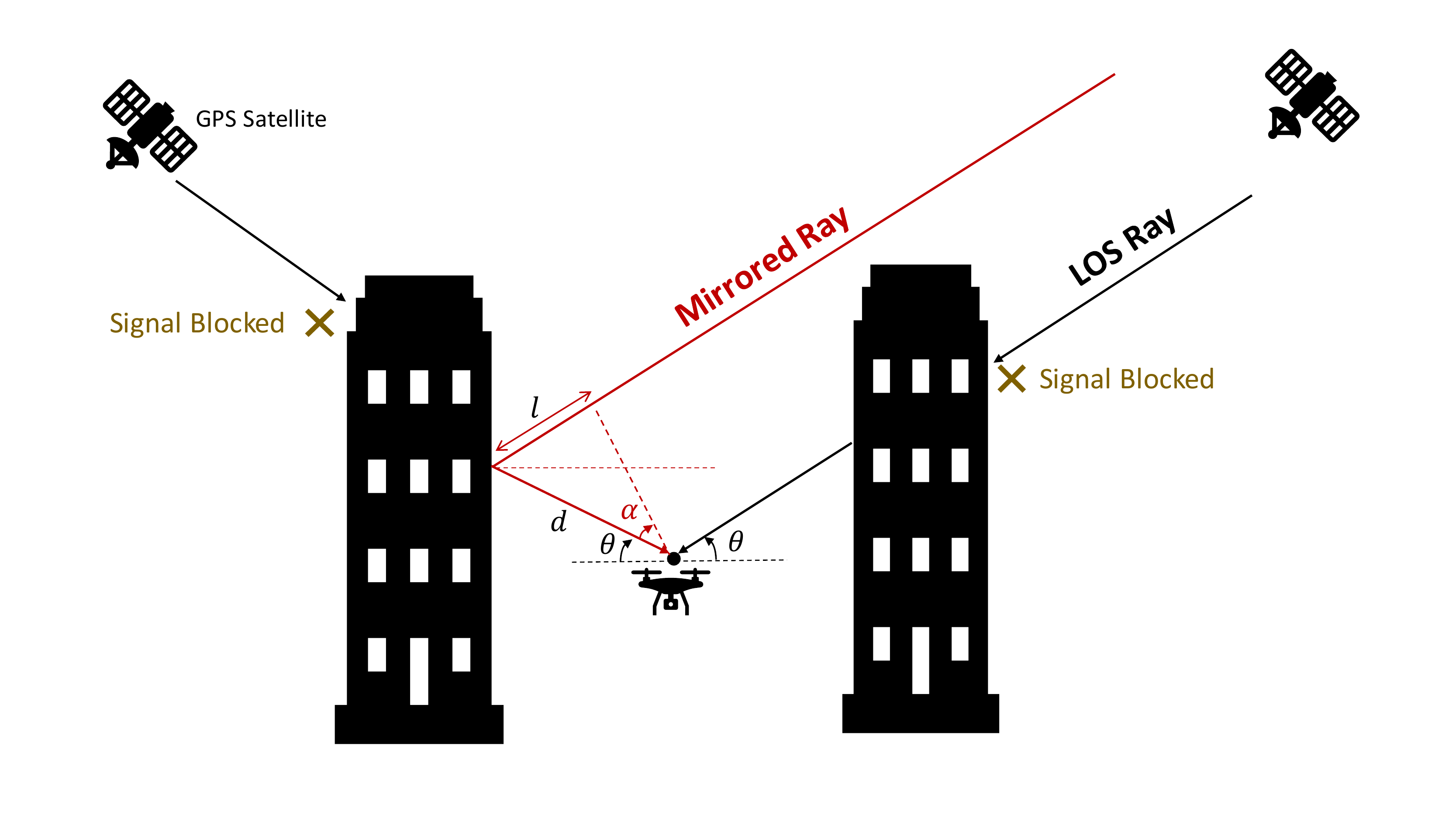}
    \caption{Multipath reflection in a typical urban setting}
    \label{fig:multipath}
\end{figure}

In this work, only a single reflection of the ray from a building's surface is considered. This is because attenuation and degradation in the GNSS signal strength due to multiple reflections causes less significant positioning error as stated in \cite{ray1999mitigation}. In addition, it is assumed that the reflected surface is normal to the ray. This is a simplification, but it captures many behaviors of GPS multipath error without detailed information on the reflection geometry and reflection surface composition. In practice, multipath reflections with high incidence angles are often ignored due to outlier rejection. Therefore, normal reflections are often the most detrimental to the performance of the system, and with the assumption of a single reflection, normal incidence angle, and clear LOS of the reflected signal we derive the 2D geometric relationship shown in \Cref{fig:multipath}.

We use the following trigonometric identity to derive the additional path length covered by the GNSS signals after reflection.
\begin{equation}
    \begin{gathered}
    \label{eq:trig}
        m = d + l \\
        m = d (1 + sin\alpha)\\
        m = d (1+sin(\pi/2 - 2\theta))
    \end{gathered}
\end{equation}

\subsection{Measurement Noise}
The pseudo ranges obtained from the GNSS constellations contain inherent randomness which can be modeled as a combination of white noise, pink noise, and random walk. Various noise models have been studied in the literature to model GNSS measurement noise. \cite{martinOUProcess,soundyOUProcess} highlight that correlated noise models fit the GNSS measurement noise better than uncorrelated noise models. It also highlighted that Ornstein Ulenback (OU) process can be utilized to model correlated GNSS noise. An OU process has the following attractive properties: mean-reverting, continuous and has few parameters; making it suitable for fast embedded computation and sensor fusion applications.   

The OU process is defined by the stochastic differential equation:
\begin{equation}
    dx_{t} = \theta_{ou}(\mu_{ou} - x_{t}) + \sigma_{ou} dW_{t} 
\end{equation}
where $\theta_{ou}$ and $\sigma_{ou}$ are parameters that can be tuned to match the randomness corresponding to a specific model of GNSS receiver and $\mu_{ou}$ is a parameter that can be adjusted to introduce drifting behavior of a position fix when the receiver velocity is slow or losing LOS to too many satellites. In this work, we use the OU process to model the measurement noise by the GNSS receiver.

\subsection{Localization}
The pseudoranges calculated by the GNSS receiver describe the distance between the satellites and the receiver as well as the errors from external sources, for example, atmospheric delays and multipath interference. To imitate this behavior in simulation, the true ranges are first computed using the true positions of the virtual satellites and the true position of the receiver using Gazebo/MoCap depending on the simulation/mixed reality test setup. For each satellite $i$, the pseudorange $P_i$ is calculated as:
\begin{equation}
\label{eq:pseudo_range}
    P_i = [(x_i - x)^2 + (y_i - y)^2 +(z_i - z)^2]^{\frac{1}{2}} + m_i + c_0 \Delta t + e
\end{equation}
The right side of (\ref{eq:pseudo_range}) is the sum of (i) mean-squared difference between the $i^{th}$ satellite's position  $(x_i,y_i,z_i)$ and the receiver's position
$(x,y,z)$, (ii) multipath offset $m_i$ calculated by the ray tracing algorithm, and (iii) additional delay terms $c_0 \Delta t$ and measurement noise, where $c_0$ represents the speed of light, $e$ accounts for measurement noise and $\Delta t$ is the delay associated with the clock offset between satellites and the receiver. 

To calculate the position of the receiver using the pseudorange measurements, a minimum of four independent pseudorange measurements are required. This is because the system of equations contains four unknown variables $(x,y,z,\Delta t)$. When the number of measurements is more than four, the system of equations becomes over-determined. To decode the position of the receiver, first (\ref{eq:pseudo_range}) is linearized about some initial estimate or guess for the receiver's position (the linearization point) using Taylor's theorem. Then, the corrections to these initial estimates are recursively adjusted to obtain the receiver's actual position and clock offset \cite{parkinson1996global}. The compact representation of the set of the linearized equation in matrix form can be described as:
\begin{equation}
\label{eq:linearize}
    \Delta P = A \Delta x +  e
\end{equation}
 in which $\Delta P$ is a vector of difference between the corrected pseudorange and modeled pseudorange of size $n$, where $n$ is the number of pseudorange measurements, $\Delta x$ designates the four-element vector of unknowns, the receiver's position, and the clock offset, from the linearization point. A is an $n\times 4$ matrix of the partial derivatives of the pseudoranges $wrt$ the unknowns and is defined as:
 \begin{equation}
\label{eq:Amatrix}
    A =  
    \begin{bmatrix}
    \frac{x_1 - x}{\rho_1} &\frac{y_1 - y}{\rho_1}&\frac{z_1 - z}{\rho_1} & -1\\
    \vdots &\vdots&\vdots & \vdots\\
    \frac{x_i - x}{\rho_i} &\frac{y_i - y}{\rho_i}&\frac{z_i - z}{\rho_i} & -1\\
    \vdots &\vdots&\vdots & \vdots\\
    \frac{x_n - x}{\rho_n} &\frac{y_n - y}{\rho_n}&\frac{z_n - z}{\rho_n} & -1\\
    \end{bmatrix}
\end{equation}
where $\rho_i = [(x_i - x)^2 + (y_i - y)^2 +(z_i - z)^2]^{\frac{1}{2}}$ is the actual distance between $i^{th}$ satellite and the receiver. The receiver then solves the matrix equation (\ref{eq:linearize}) by minimizing the estimation error using the least squares algorithm to get,
\begin{equation}
\label{eq:least_squares}
    \Delta x = (A^{T}A)^{-1}A^{T} \Delta P
\end{equation}
In general, when the initial estimates are incorrect, the solution of the non-linear problem requires multiple iterations to converge. The least squares algorithm is iterated until $\Delta x$ becomes close to a small value $\epsilon$, which in practice is kept typically around $10^{-6}$.

\subsection{Localization Accuracy}
As the measurements obtained from the satellites are not perfect, due to noise and other unmodeled errors, the position estimates given by (\ref{eq:least_squares}) are also not accurate. Assuming (a priori) standard deviation of modeling error in pseudorange measurements, $\sigma$, the covariance matrix $Q$ of the expected error in unknown parameters $(x,y,z,\Delta t)$ can be formulated as:
\begin{equation}
\label{eq:covariance}
    Q = (A^{T}A)^{-1} \sigma^2 \equiv D \sigma^2  
\end{equation}
With a given standard deviation of measurement errors, $\sigma$, the quality of least squares solutions can be computed by using the diagonal elements of the $Q$ matrix described as:
\begin{equation}
\label{eq:qmatrix}
    Q = \begin{bmatrix}
    {\sigma_x}^2 &\sigma_{xy}&\sigma_{xz} & \sigma_{xt}\\
    \sigma_{yx} &{\sigma_y}^2&\sigma_{yz} & \sigma_{yt}\\
    \sigma_{zx} &\sigma_{zy}&{\sigma_z}^2 & \sigma_{zt}\\
    \sigma_{tx} &\sigma_{ty}&\sigma_{tz} & {\sigma_t}^2\\
    \end{bmatrix}
\end{equation}
where $\sigma_{x},\sigma_{y}, \sigma_{z}$ are the estimated \textit{rms} errors in the user position $(x,y,z)$ and $\sigma_{t}$ is the \textit{rms} value of the error in the user clock bias estimate. $\sigma_{xy},\sigma_{yz}, \sigma_{zx}$ are covariance in the position estimates in $x$, $y$ and $z$ directions, and $\sigma_{xt},\sigma_{yt}, \sigma_{zt}$ are the covariance of position estimates $(x,y,z)$  with the clock bias $\Delta t$. Observe that the elements of matrix $Q$ are dependent only on the receiver-satellite geometry. The overall quality of the least-squares solution can be obtained by taking the square root of the diagonal elements of the $Q$ matrix.
\begin{align}
\label{eq:gdop}
    \sigma_G &= \sqrt{{\sigma_x}^2+{\sigma_y}^2+{\sigma_z}^2+{\sigma_t}^2}\\ 
    &=\sqrt{{D_{11}}^2+{D_{22}}^2+{D_{33}}^2+{D_{44}}^2} \sigma
\end{align}

\begin{figure}[H]
    \centering
    \begin{subfigure}{0.45\textwidth}
        \centering
        \includegraphics[width=0.5\textwidth]{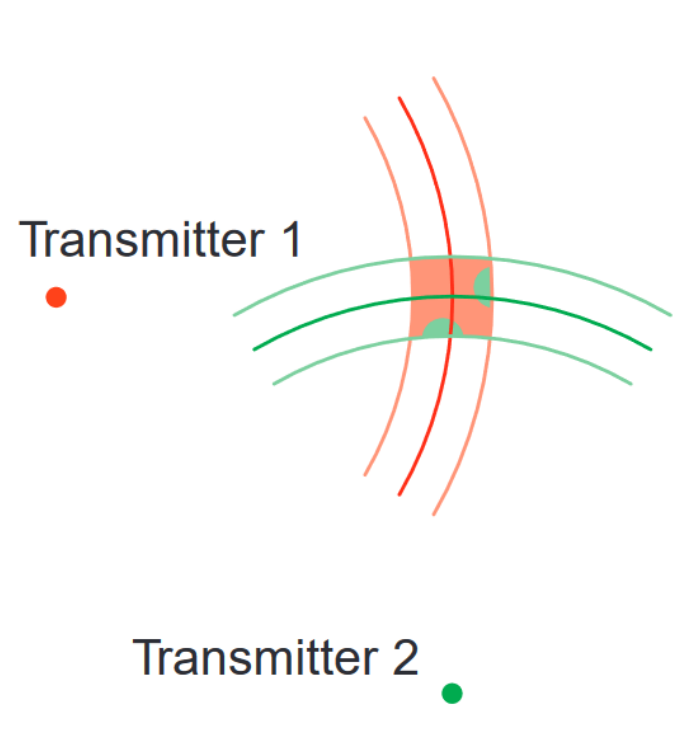}
        \caption{Small uncertainty when transmitters are far away}
        \label{fig:dop_a}
    \end{subfigure}
    \begin{subfigure}{0.45\textwidth}
        \centering
        \includegraphics[width=0.5\textwidth]{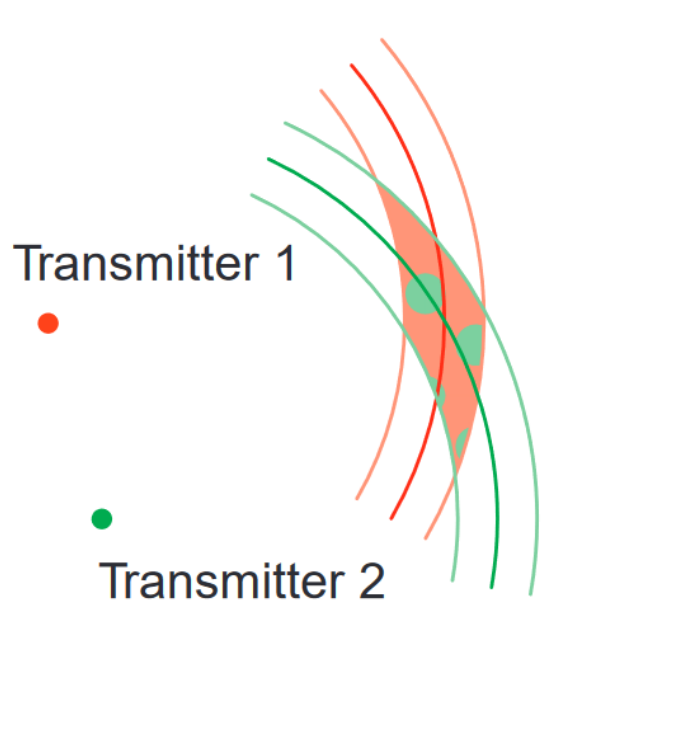}
        \caption{Large uncertainty when transmitters are closer}
        \label{fig:dop_b}
    \end{subfigure}
    \caption{Influence of satellite-receiver geometry on GNSS position accuracy adapted from \cite{Langley1999DilutionOP}}
    \label{fig:dop}
\end{figure}

The standard deviation, $\sigma_G$, is a scaled version of the standard deviation of the measurement errors. As observed by authors in \cite{isik2020gnssintegrity}, the scaling factor is typically greater than 1 which results in the standard deviation of the least squares solution being greater than $\sigma$. This inflation of variance of the estimated unknown variables due to receiver-satellite geometry is known as Dilution of Precision (DOP) and $\sigma_G$ is called geometric dilution of precision (GDOP). To understand this effect better, let us consider two satellites transmitting to a ground receiver in two different configurations as depicted in \Cref{fig:dop}. The receiver's true position can be uniquely identified by intersecting the two range circles. However, due to uncertainty in the measurements, the range circles are inexact, resulting in an error in positioning. This
error depends on the geometry related to the receiver and the transmitters. When the transmitters are located far apart shown in \Cref{fig:dop_a}, there is a small region of uncertainty where the receiver must be located, whereas, the region of uncertainty is larger when the transmitters are closer to each other as shown in \Cref{fig:dop_b}. 
Instead of assessing the quality of the overall solution, typically the specific components of DOPs, i.e., the three-dimensional position component (PDOP), the horizontal component (HDOP), the vertical component (VDOP), and the time component (TDOP) are evaluated. Each of these components is computed as:
\begin{align}
\label{eq:dop}
    \text{PDOP} &=\frac{\sqrt{{\sigma_x}^2+{\sigma_y}^2+{\sigma_z}^2}}{\sigma} \\ 
    \text{HDOP}&=\frac{\sqrt{{\sigma_x}^2+{\sigma_y}^2}}{\sigma} \\
    \text{VDOP} &= \frac{\sigma_z}{\sigma}\\ 
    \text{TDOP}&= \frac{\sigma_t}{\sigma}
\end{align}
Note that $\text{PDOP}^2 = \text{HDOP}^2 + \text{VDOP}^2$ and  $\text{GDOP}^2 = \text{PDOP}^2 + \text{TDOP}^2$. The DOP values can be pre-computed as they are only dependent on the receiver and satellite coordinates. The higher the DOP values the poorer the accuracy of the GNSS position fix. Although, there is no common agreement on the classification of the quality of the GNSS solution based on DOP coefficient values. Based on the study conducted in \cite{isik2020gnssintegrity}, Table. \ref{tab:DOPvalues} highlights the rating of the resulting solution as per the DOP values.   
\begin{table}[H]
    \caption{Characterization based on DOP coefficient}
    \label{tab:DOPvalues}
    \renewcommand{\arraystretch}{1.2}
    \centering
    \begin{tabular}{p{2.5cm}p{1.5cm}p{1.5cm}p{1.5cm}p{1.5cm}p{1.5cm}p{1.5cm}}
        \hline
        \textbf{DOP value} & $\textbf{<1}$ & $\textbf{1-2}$ & $\textbf{2-5}$ & $\textbf{5-10}$ & $\textbf{10-20}$ & $\textbf{>20}$ \\
        \hline
         \textbf{Rating} & Ideal & Exellent & Good  & Moderate & Fair & Poor \\
        \hline
    \end{tabular}
    
\end{table}

%% file: sections/04_architecture.tex

\section{Architectural Design}
\label{sec:arch-soft-design}
\vspace{2pt}
\begin{figure}[t]
    \centering
    \includegraphics[width=\textwidth]{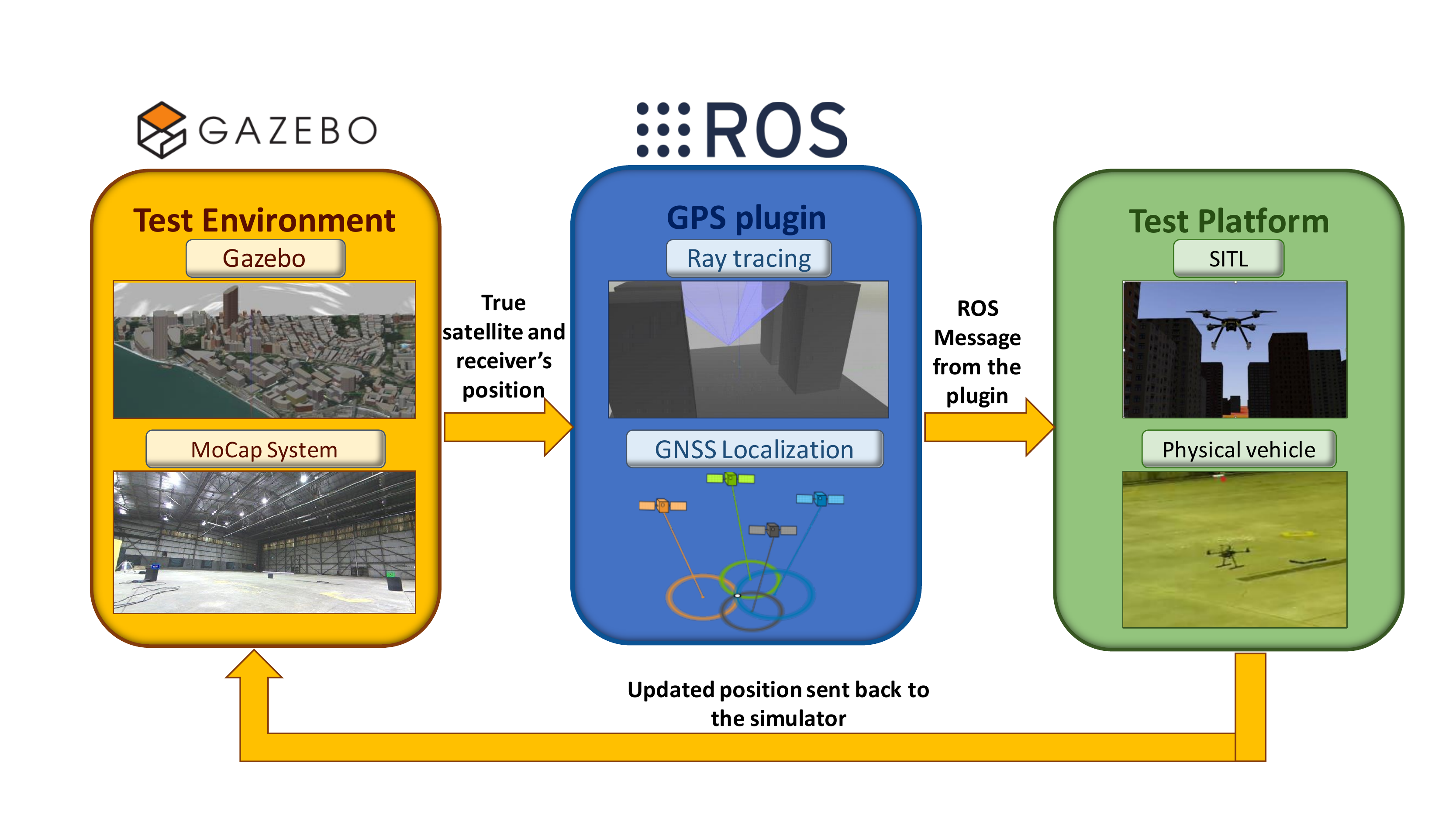}
    \caption{GNSS multipath plugin flowchart}
    \label{fig:flowchart}
\end{figure}
Before describing the architecture of the proposed plugin, it is important to understand some useful terminologies related to ROS and Gazebo. This section explains the basic tools and terminologies used for designing Gazebo plugins and presents the software architecture of the proposed GNSS multipath plugin.  
\subsection{ROS and Gazebo}
The Robot Operating System(ROS) is a framework for the development and deployment of robotic systems \cite{Quigley09}. It contains collections  of drivers, libraries, and tools for interprocess communication between various robot hardware and software. ROS follows a computational graph structure in which ROS processes are graph nodes. The nodes transfer information among each other using edges called \textit{ROS topics}. Each ROS topic broadcasts information in the form of \textit{ROS messages}. This inter-process communication allows easy development of user libraries and executables.    

Gazebo is a 3D simulator built for both indoor and outdoor robotic applications \cite{Gazebo}. Together with ROS, Gazebo provides an efficient physics platform for high fidelity 3D simulation of various robotic systems. The interface between ROS and Gazebo is controlled by a set of shared libraries called \textit{gazebo\_plugins}. 
\subsection{Gazebo Plugins}
A plugin in Gazebo is a shared library that can access all the functionalities of the simulator from the physics engine to the simulated environment. The plugins are flexible pieces of code that can be easily added and removed from the existing simulation system. There are four different types of Gazebo plugins to modify and control properties of various simulated entities:
\begin{enumerate}
    \item {\textbf{World Plugin}: for altering physics settings.}
    \item {\textbf{Sensor Plugin}: for changing sensor parameters.}
    \item {\textbf{Visual Plugin}: for modifying visual properties of the 3D models.}
    \item {\textbf{System Plugin}: for system-level settings.}
\end{enumerate}

\subsection{Simulation Description File (SDF)}
Gazebo uses an XML-based configuration file for saving and loading the parameters related to the simulated models. It captures the necessary information about the \textit{scene}, \textit{physics}, \textit{lighting}, and \textit{plugins} of the Gazebo simulator.  
\subsection{GNSS Multipath Plugin}
The architecture of the GNSS plugin is similar to the different sensor plugins already present in the Gazebo simulator. The sensor plugins can be attached to any 3D object in the simulator using Gazebo's SDF file. The plugin is developed as a C++ class, which inherits from ray sensor plugin of the Gazebo. In Gazebo, the ray sensor is the backbone for LiDAR and laser-based range sensors. The line segments mentioned in the previous section are constructed by orienting rays from the receiver in the direction of the satellites (azimuth and elevation angles). Using the native C++ APIs of the ray sensor plugin, the additional pseudorange offset is calculated by using algorithm \ref{alg:multipatherror}. Our major contribution is to modify and optimize the existing ray sensor plugin APIs to allow real-time updating of the azimuth and elevation angles of the satellites. The noise levels and satellite constellations can be modified using the SDF, allowing the proposed GNSS plugin to simulate a wide range of GNSS constellations and a variety of commercially available GNSS receivers.  
\begin{algorithm}[h]
    \caption{GNSS Localization}
    \label{alg:receiverpos}
    \KwIn{True Ranges $\rho_i$, satellites position $(x_{sat}, y_{sat}, z_{sat})$, elevation angles $\theta_i$, multipath offset $m_i$, measurement noise $e_i$  }
    \KwOut{Receiver's position, ($x,y,z$)}
    \For{$i\in (1,n)$}{
        \tcp{Checking for low elevation angle}
            \If{$\theta_i > \theta_0$}{
            $m_i \gets ComputeMultipathOffset(r_i^{LOS}, r_i^{REF})$ \tcp{ multipath offset, $m_i$, is computed by the ray tracing algorithm}
            $P_i \gets \rho_i + m_i + e_i$ \tcp{$P_i$ is the computed pseudorange, $e_i$ is the measurement error}  
            $vis\_sat \gets vis\_sat + 1$ \tcp{visible satellites are incremented} 
            }
        }
        \tcp{Checking for number of visible satellites}
        \If{$vis\_sat>4$}{
            $(x,y,z) \gets RecursiveLeastSquares(P, (x_0, y_0,z_0))$ \tcp{receiver's position, $(x,y,z)$, calculated using RLS algorithm, $(x_0,y_0,z_0)$ is the initial guess of the receiver's position}
            $DOP\_coeff \gets ComputeDOP((x_{sat}, y_{sat}, z_{sat}), (x,y,z))$ \tcp{compute DOP coefficient using satellites and receiver's estimated position}
        }
\end{algorithm}
\begin{algorithm}[ht]
    \caption{Multipath Offset Calculation}
    \label{alg:multipatherror}
    \KwIn{Ranges $r_i^{LOS}$ and $r_i^{REF}$, $i \in (1, n)$ from the ray sensor. where n is the number of satellites, $r_i^{LOS}$ is the laser range for LOS ray and $r_i^{REF}$ is the laser range for mirror ray.}
    \KwOut{Multipath offset, $m_i$}
    \For{$i\in (1,n)$}{
        \tcp{Checking for LOS obstruction}
        \If{$r_i^{LOS} < MAX\_RANGE$}{
            \tcp{Checking whether the mirror ray exists}
            \If{$r_i^{REF} < MAX\_RANGE$}{
                $m_i \gets CalculateOffset(r_i^{REF})$\tcp{using trigonometric identities to calculate pseudorange offset}
            }    
        }
        \tcp{No GPS multipath}
    }
\end{algorithm}

Algorithm. \ref{alg:receiverpos} describes the estimation of the receiver's position and calculation of the estimation accuracy. For calculating the input ranges, $\rho$, the true position of the receiver is extracted from the simulator/MoCap system based on the simulation/mixed reality experiment. The plugin obtains the position of each satellite using the ephemeris data provided. The true range for each satellite is computed using the extracted position of the receiver and the satellite. Using Algorithm. \ref{alg:multipatherror}, the plugin computes the multipath offset, $m_i$, for each satellite. The low elevation satellites are excluded from the estimation by checking against a threshold angle, $\theta_0$, as those are most prone to atmospheric effects and multipath errors \cite{moller2019atmospheric}. Finally, the coordinates of the receiver are computed only if the minimum number of visible satellites is more than 4. The recursive least square estimation algorithm is used to estimate the receiver's coordinates. The various components of the DOP (PDOP, HDOP, VDOP, and TDOP) are computed using the receiver and the satellite's positions.

The multipath pseudorange offset is computed using the ray tracing algorithm defined in Algorithm. \ref{alg:multipatherror}. The ray sensor first calculates the ranges of line-of-sight and reflected rays, $r_i^{LOS}$ and $r_i^{REF}$, to infer the presence of multipath errors. Multipath errors are detected when the line-of-sight ray is obstructed and there exists a obstructed reflected ray (indicating the presence of a building or a reflecting surface). If the range criterion is satisfied, the function \textit{calculateOffset} computes the additional pseudorange offsets using the derived trigonometric identities in (\ref{eq:trig}). In the absence of direct line of sight to the satellite (LOS) and lack of a building for reflection, the satellite is declared as not visible (no signal received).  

The plugin logs the computed coordinates of the receiver, components of the DOP, visible satellites, noise levels, and the pseudorange offset of the satellites. All this information is wrapped into a custom ROS message and published as a ROS topic for downstream applications such as path planning and navigation.

\subsection{GNSS Multipath Plugin's message and SDF}
As mentioned previously, the plugin combines all the information into a ROS message and  publishes it using a ROS topic. The details of each entry of the ROS message are tabulated in Table. \ref{tab:ROSmsg}.
\begin{table}[H]
    \renewcommand{\arraystretch}{1.2}
    \centering
    \begin{tabular}{p{2.5cm}p{4cm}p{7.0cm}}
        \hline
        \textbf{Field} & \textbf{Datatype} & \textbf{Purpose}  \\
        \hline
        \texttt{header} & \texttt{std\_msgs/Header} & Stores the timestamp and frame information\\
        \texttt{rec\_pos} & \texttt{sensor\_msgs/NavSatFix} & Stores the receiver's geodetic location\\
        \texttt{dop} & \texttt{float32[]} & Stores the components of Dilution-of-Precision\\
        \texttt{range\_offset} & \texttt{float32[]} & Stores the multipath offset for each satellite[m] \\
        \texttt{sats\_blocked} & \texttt{int32[]} & Stores the PRN of blocked satellites\\
        \texttt{num\_vis\_sat} & \texttt{int32} & Stores the number of visible satellite\\
        \texttt{num\_block\_sat} & \texttt{int32} & Stores the number of blocked satellite\\
        \texttt{noise} & \texttt{float32[]} & Stores the measurement noise [m]\\
        \hline
    \end{tabular}
    \vspace{2pt}
    \caption{ROS message for GNSS multipath plugin}
    \label{tab:ROSmsg}    
\end{table}
Below is the snippet of the SDF for adding the GNSS multipath plugin. To add the plugin to an existing model these lines must be included. More details about the SDF can be found at \\ \href{https://github.com/kpant14/multipath_sim/tree/main/worlds}{\texttt{https://github.com/kpant14/multipath\_sim/tree/main/worlds}}
\lstset{language=XML}
\begin{lstlisting}
<plugin name="laserscan" filename="libMultipathSensorPlugin.so">
    <ros>
        <namespace>/multipath</namespace>
        <remapping>~/out:=scan</remapping>
    </ros>
    <output_type>sensor_msgs/LaserScan</output_type>
    <disableNoise>0</disableNoise>
    <satNum>8</satNum>
    <satAzimuth>2.93 3.39 5.56 0.23 4.02 1.38 2.23 0.65</satAzimuth>
    <satElevation>0.78 0.32 1.02 0.90 0.47 0.26 1.09 0.38</satElevation>
</plugin>
\end{lstlisting}

%% file: sections/05_experimental_results.tex
\section{Experimental Results}
\label{sec:impl}
\vspace{2pt}
In this section, we examine the performance of the proposed plugin by comparing its performance with the publicly available Urban-Nav dataset, containing the GNSS records of the city of Hong Kong.  
\subsection{Urban-Nav Dataset}
\begin{figure}[H]
    \centering
    \begin{subfigure}{0.4\textwidth}
        \centering
        \includegraphics[width=\textwidth]{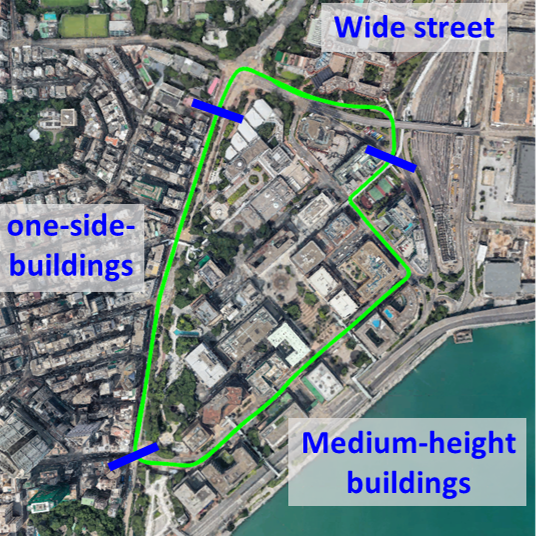}
        \caption{Map of Hong Kong where the data is recorded.}
        \label{fig:urbannav_a}
    \end{subfigure}
    \begin{subfigure}{0.5\textwidth}
        \centering
        \includegraphics[width=\textwidth]{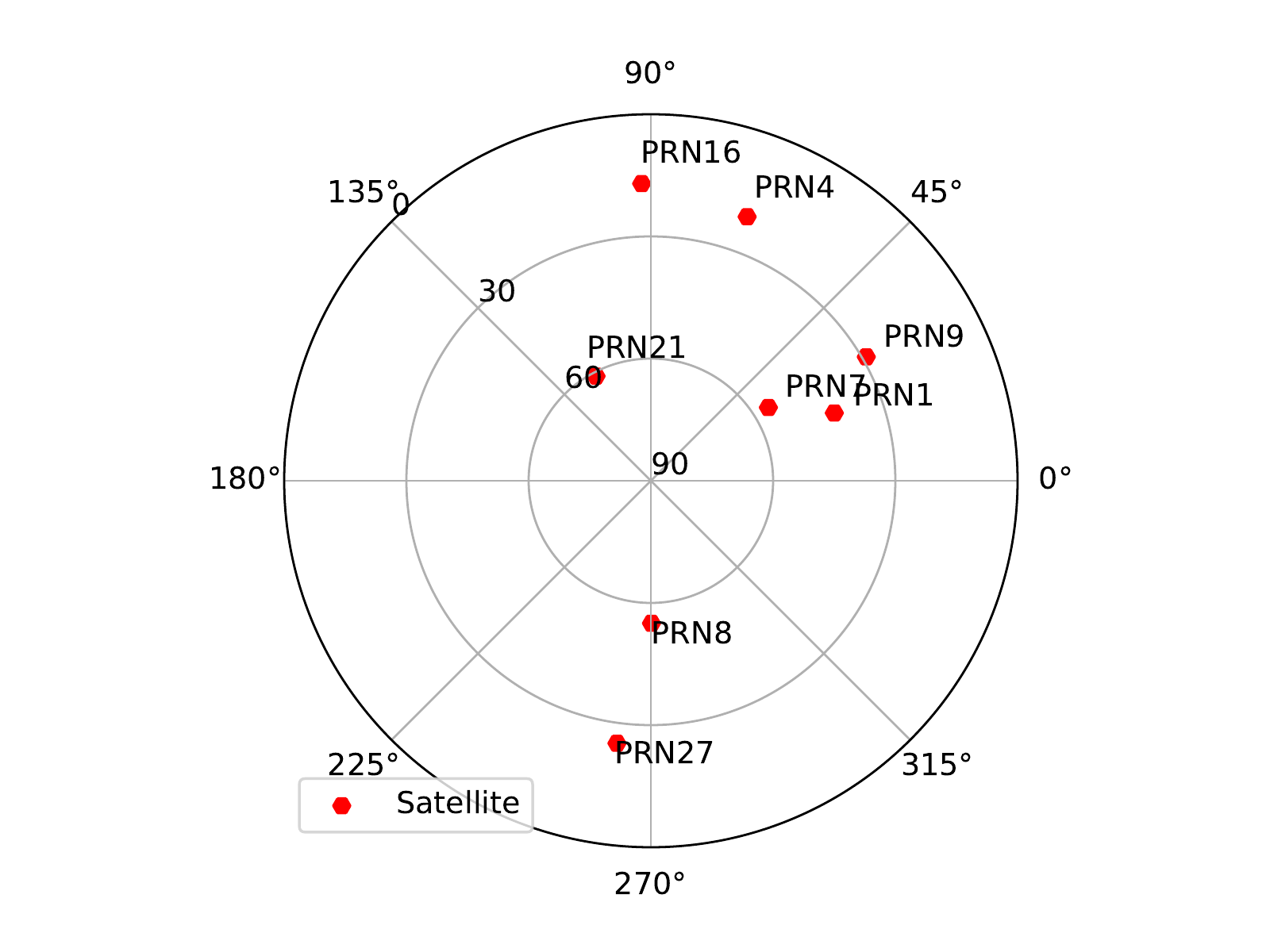}
        \caption{Satellite constellation diagram for u-blox EVK-M8T}
        \label{fig:urbannav_b}
    \end{subfigure}
    \caption{Urban-Nav dataset for Hong Kong city: Map and Satellite Constellation (adapted from  Hsu et. al. \cite{HsuUrbannav})}
    \label{fig:urbannav}
\end{figure}

We utilize the open-source Urban-Nav Dataset collected and prepared by Hsu et. al. \cite{HsuUrbannav} to test the simulation of multipath error from our plugin. The dataset contains multiple GNSS navigation records of the city of Hong Kong and Tokyo in the form of RINEX files measured using various GNSS receivers, namely, u-blox ZED-F9P, EVK-M8T and NovAtel Flexpak6, etc. The ground truth position of the receiver is computed using sensor fusion from IMU and NovAtel SPAN-CPT receiver. \Cref{fig:urbannav_a} shows the map of Hong Kong where the data is acquired which we used in our study. The recorded data captures the essential components of an urban environment, for example, medium and tall buildings and wide streets. \Cref{fig:urbannav_b} illustrates the satellite constellation diagram at the begining of the data acquisition.  

\subsection{Results}
\begin{figure}[H]
    \centering
    \includegraphics[width =\textwidth]{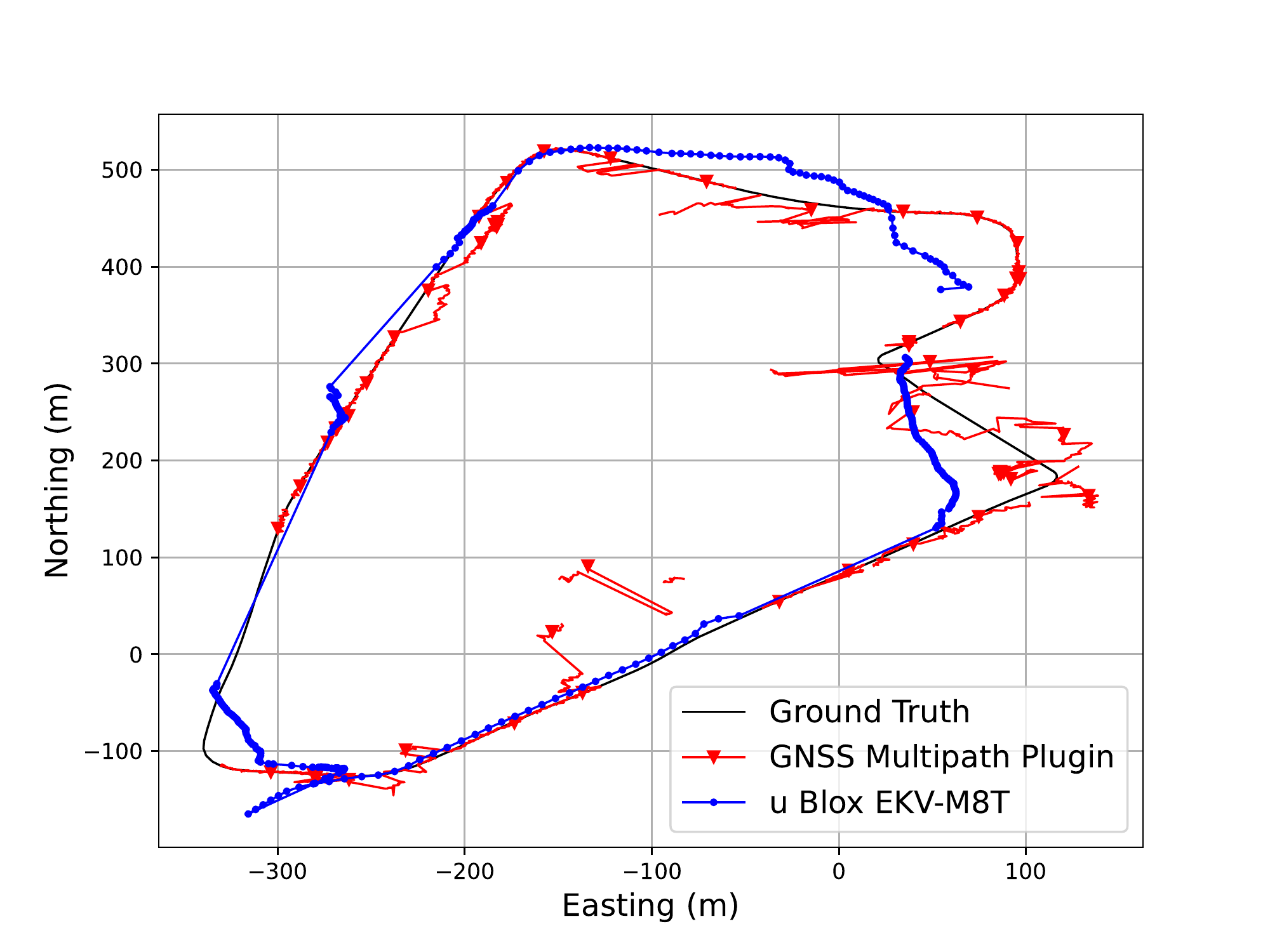}
    \caption{Comparison of GNSS position fix of u-blox EVK-M8T and the proposed plugin}
    \label{fig:comparison}
\end{figure}

In \Cref{fig:comparison}, the GNSS position solution obtained from the plugin is depicted. We observe a close match between the noise properties of the simulated position fix by the proposed plugin and the measured GNSS position solution by a commercially available receiver, u-blox EKV-M8T. The exact position fix differ as the commercially available receivers performs additional post-processing on the raw position fix generated by the recursive least squares estimation algorithm by using an Extended Kalman Filter (EKF). Although there is a mismatch in the position fix, the proposed plugin is able to effectively capture the multipath prone regions in the city. The plugin is also able to capture the loss of position fix during the phase of the trajectory when vehicle passes near tall buildings. This is attributed to the obstruction of the satellites due to the buildings.

\begin{figure}[H]
    \centering
    \includegraphics[width =0.9\textwidth]{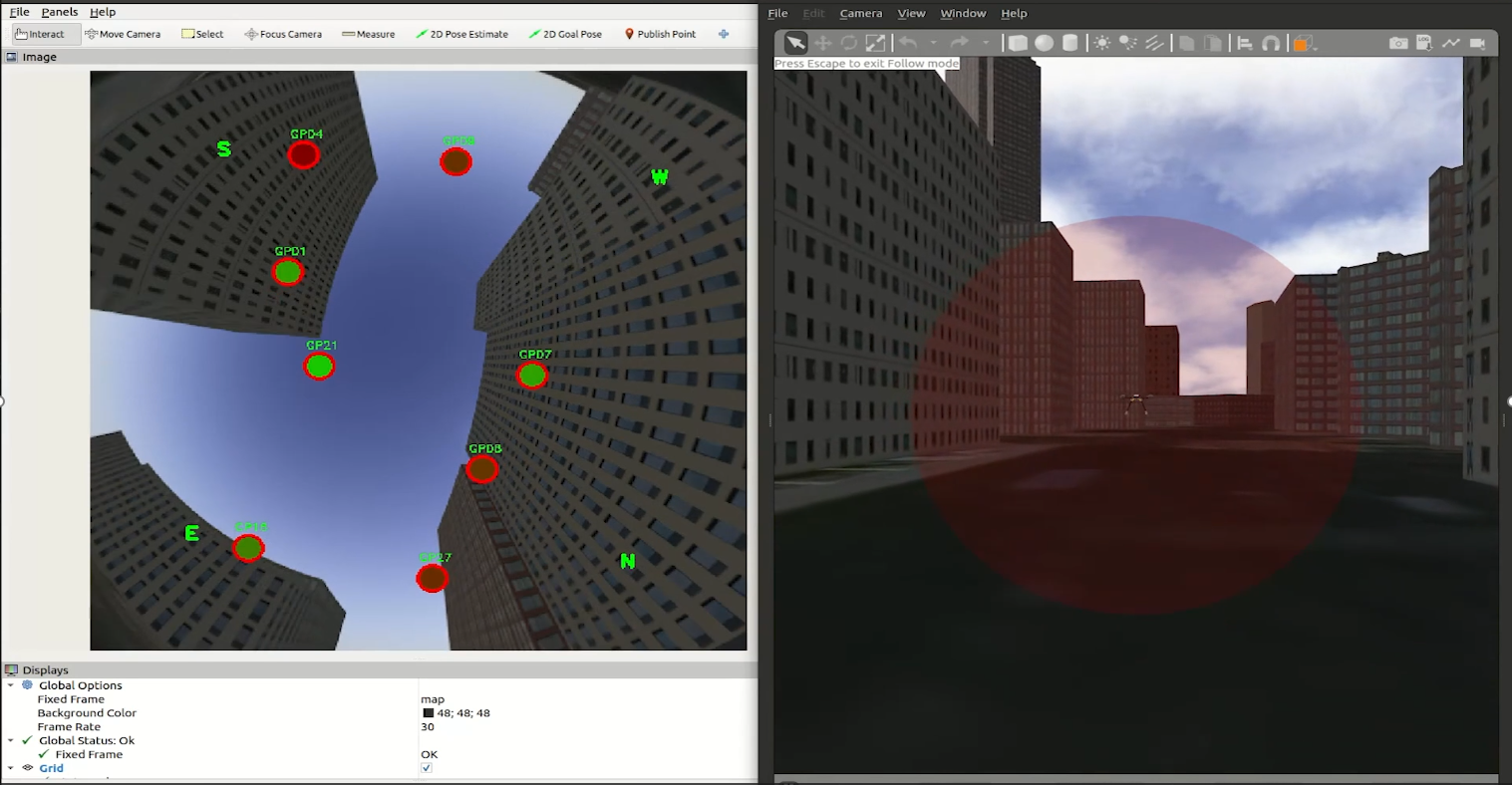}
    \caption{An instance of the GNSS multipath plugin being attached to a drone model in the Gazebo simulator}
    \label{fig:plugin_instance}
\end{figure}

\begin{figure}[H]
    \centering
    \includegraphics[width=0.8\textwidth]{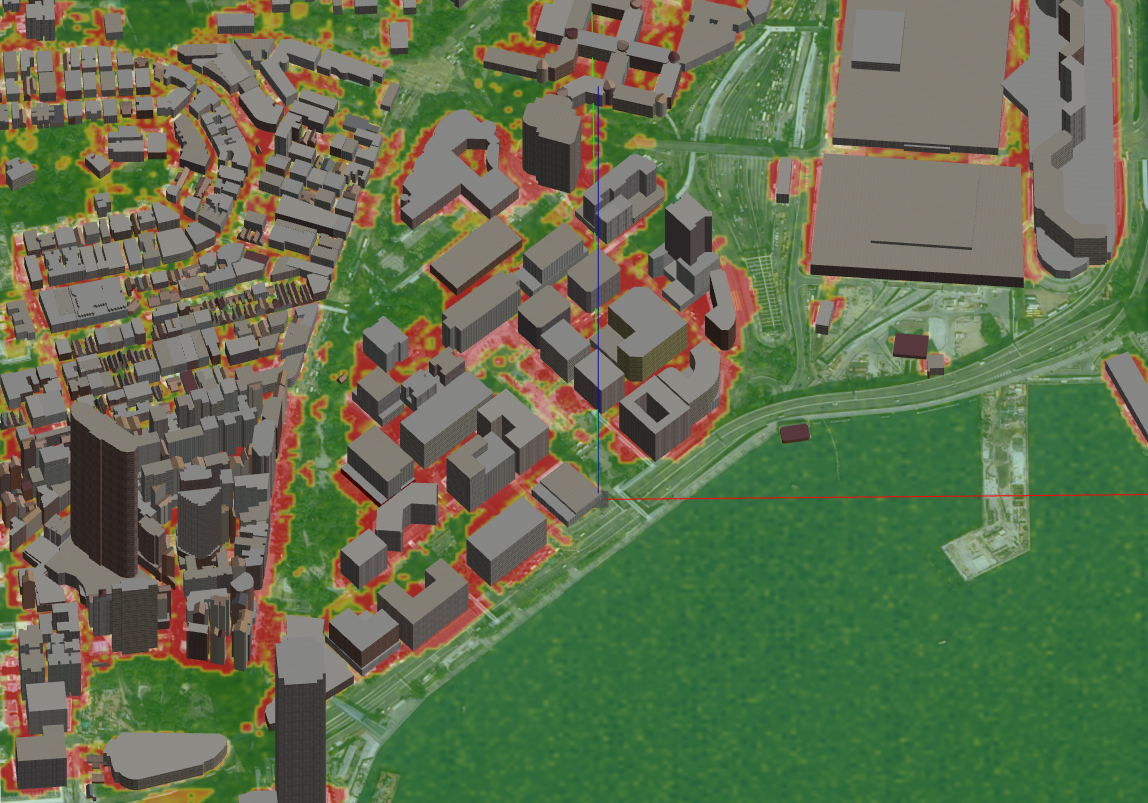}
    \caption{Heat Map depicting multipath error in virtual Hong Kong environment}
    \label{fig:heatmap}
\end{figure}

\Cref{fig:plugin_instance} shows a typical instance of the plugin, the left hand side shows the sky plot using a simulated fish-eye camera pointing up toward the sky, and, on the right side a drone is equipped with our proposed plugin. The spherical bubble around the drone shows the intensity of the multipath error at the given location. Additionally, we have created a heat map highlighting the intensity of the GNSS position error throughout the entire city at an altitude of 15m as shown in \Cref{fig:heatmap} to understand the effects of the multipath errors. 


%% file: sections/06_conclusion.tex
\section{Conclusion and Future Work}
\label{sec:conclusion}
\vspace{2pt}
In this paper, we presented the design of a Gazebo plugin for GNSS sensor in a virtual 3D environment. Our design enables the construction of a virtual 3D environment of a part of any city and presents a realistic high-fidelity simulation of various commercially available GNSS receivers. We chose the ROS-Gazebo framework because of its ubiquitous application and support in the robotics community. The proposed plugin has a wide range of applications, especially in the growing field of high-assurance robotics and autonomy.
In the future, we plan to extend the functionality of the proposed GNSS plugin to incorporate the orbital equations of the satellites to ease the process of retrieving the coordinates of the satellites in a constellation. Finally, we plan to officially release our plugin inside the Gazebo community.